\documentclass{article}
\usepackage{iclr2027_conference,times}

\usepackage{amsmath,amssymb}
\usepackage{booktabs}
\usepackage{graphicx}
\usepackage{multirow}
\usepackage[table]{xcolor}
\usepackage{tabularx}
\usepackage{wrapfig}
\definecolor{opdrow}{RGB}{250,236,232}
\newcommand{\oc}{\cellcolor{opdrow}}
\definecolor{ablrow}{RGB}{234,245,236}
\newcommand{\gc}{\cellcolor{ablrow}}
\newcolumntype{C}{>{\centering\arraybackslash}X}

\usepackage{amsmath,amsfonts,bm}

\def\eqref#1{equation~\ref{#1}}

\def\1{\bm{1}}

\DeclareMathAlphabet{\mathsfit}{\encodingdefault}{\sfdefault}{m}{sl}
\SetMathAlphabet{\mathsfit}{bold}{\encodingdefault}{\sfdefault}{bx}{n}

\usepackage{hyperref}
\usepackage{url}

\newcommand{\pS}{\pi_{\theta}}
\newcommand{\pT}{\pi_{\mathrm{T}}}
\newcommand{\best}[1]{\textbf{#1}}

\title{Train Where the Quantized Model Goes:\\ On-Policy Distillation for Low-Bit Reasoning}

\author{Yuanteng Chen}  

\iclrfinalcopy

\newcommand{\arxivnameline}{%
  Yuanteng Chen\textsuperscript{1,2,3,*},
  Zhilei Liu\textsuperscript{1,2,*},
  Peisong Wang\textsuperscript{1,2,\textdagger},
  Yuantian Shao\textsuperscript{1,5},
  Chuangyi Li\textsuperscript{1,2},\\[2pt]
  Weining Wang\textsuperscript{1,2,3},
  Shuang Qiu\textsuperscript{4},
  Gang Li\textsuperscript{1,2},
  Jing Liu\textsuperscript{1,2,3},
  Jian Cheng\textsuperscript{1,2,3,\textdagger}}

\newcommand{\arxivaffilblock}{%
  \textsuperscript{1}\,Institute of Automation, Chinese Academy of Sciences\\
  \textsuperscript{2}\,School of Artificial Intelligence, University of Chinese Academy of Sciences\\
  \textsuperscript{3}\,Zhongguancun Academy
  \quad
  \textsuperscript{4}\,City University of Hong Kong
  \quad
  \textsuperscript{5}\,NJUST\\
  \textsuperscript{*}\,Equal contribution
  \quad
  \textsuperscript{\textdagger}\,Corresponding authors}

\makeatletter
\def\@maketitle{\vbox{\hsize\textwidth
\parskip=0pt
{\LARGE\sc \@title\par}
\vskip 10pt
{\centering\normalsize\bfseries\arxivnameline\par}
\vskip 7pt
{\centering\normalsize\arxivaffilblock\par}
\vskip 0.2in minus 0.05in}}
\makeatother

\begin{document}

\maketitle

\begin{abstract}
Quantization-aware distillation (QAD) restores much of the short-form
question-answering performance lost to sub-3-bit quantization, yet leaves
mathematical and code reasoning substantially impaired. Long generations often
degenerate into repetitive loops, exhausting the decoding budget without
completing a solution. We trace this gap to quantization-amplified exposure
bias: QAD trains on fixed corpus prefixes, while quantization-induced
deviations compound along the model's own autoregressive trajectories. To
address this mismatch, we introduce an on-policy distillation (OPD) stage that
places teacher supervision where the quantized model actually goes. Starting
from a QAD checkpoint, the student generates through the quantized forward path
used at deployment and receives feedback from a frozen full-precision teacher
on its own prefixes, combining dense token-level guidance with task-verifier
rewards. Across four models at 2.79 and 1.88 effective bits, OPD raises average
BF16 performance retention from 35\% to 70\% on MATH-500 and from 66\% to 91\%
on HumanEval while preserving short-form performance, with reasoning gains
substantially exceeding those of continued teacher-forced QAD in matched-budget
comparisons. By coupling QAD's stable low-bit initialization with OPD's
on-policy reasoning recovery, our framework provides a comprehensive sub-3-bit
solution that preserves broad capabilities while restoring long-form reasoning.
Code is available at~{\hypersetup{pdfborder={0 0 0}}\href{https://github.com/MingZwhy/QAOPD}{\raisebox{-0.18ex}{\includegraphics[height=1.05em]{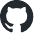}}\,GitHub}}.
\end{abstract}

\section{Introduction}

Large language models \citep{deepseekr1} impose substantial memory and
bandwidth demands, limiting inference efficiency and deployment on constrained
hardware. Quantization \citep{dettmers2022llmint8} reduces these costs by
representing model weights at lower precision while retaining the original
architecture. Pushing precision below three bits offers
substantial compression, making extreme low-bit quantization an attractive
route to deploying capable models with fewer resources.

At four bits and above, post-training quantization (PTQ)
\citep{frantar2023gptq,shao2024omniquant} often preserves
performance with only a small calibration set. Below three bits, however,
quantization errors become harder to compensate for, and direct PTQ can
severely degrade or even collapse model performance. This makes
quantization-aware training (QAT) \citep{liu2024llmqat,ma2024bitnet}
essential for adapting the model to low-bit
computation. Conventional QAT typically relies on pretraining-style data and
objectives, while modern models acquire much of their instruction-following and
reasoning capabilities during post-training. Quantization-aware distillation
(QAD) provides a practical route to recovering these capabilities: a
full-precision teacher supervises its quantized counterpart on a manageable
distillation corpus.

Our study reveals a sharp imbalance in what QAD recovers. Across four models
and two effective bit widths, QAD retains an average of 82\% of BF16
performance on short-form question answering, but only 35\% on MATH-500. Long
generations expose a striking failure mode: the quantized model enters
repetitive loops and exhausts its decoding budget without completing a
solution. The recovery deficit becomes more pronounced on tasks requiring
longer generations, making the ability to sustain and complete extended
reasoning a central target of low-bit recovery.

We trace this gap to quantization-amplified exposure bias. QAD trains the
quantized student to match its teacher on prefixes drawn from a fixed corpus.
During deployment, the student instead conditions on its own previous
predictions. Quantization perturbs the next-token distribution at every step,
and each departure changes the context for the predictions that follow. Even a
locally plausible continuation can move the model away from the trajectories
covered during training. Over a long generation, these deviations accumulate,
exposing the student to states on which it has received little supervision. The
missing guidance therefore lies along the reasoning trajectories the quantized
model actually generates.

We address this gap with a two-stage recovery framework that follows QAD
initialization with on-policy distillation (OPD). QAD first restores broad
capabilities and provides a viable low-bit policy. OPD then places teacher
supervision on that policy's own trajectories. The student samples completions
through the quantized forward path used at deployment, and a frozen BF16
teacher provides token-level feedback on the prefixes it produces. Generating
through the low-bit path makes quantization-induced changes in the student's
behavior part of the training distribution itself. Task verifiers supply
complementary rewards for final-answer correctness in mathematics and
successful test execution in code. By shifting teacher supervision from fixed
corpus prefixes to the quantized student's own trajectories, OPD targets the
accumulated deviations that disrupt long-form reasoning.

We evaluate this framework on Qwen3-0.6B, Qwen3-1.7B, Qwen3-4B, and Falcon3-1B
at 2.79 and 1.88 effective bits. Building on QAD initialization, OPD raises
average BF16 retention from 35\% to 70\% on MATH-500 and from 66\% to 91\% on
HumanEval while preserving short-form performance, with reasoning gains
substantially exceeding those of continued teacher-forced QAD under matched
corpora and optimizer-step budgets. OPD accounts for a larger share of recovery
at lower bit widths, making on-policy recovery increasingly valuable under
aggressive quantization. With only a few hundred additional optimizer steps,
the combined pipeline couples QAD's stable low-bit initialization with OPD's
reasoning recovery to preserve broad capabilities and restore long-form
reasoning below three bits.

\begin{figure}[t]
\centering
\includegraphics[width=\linewidth]{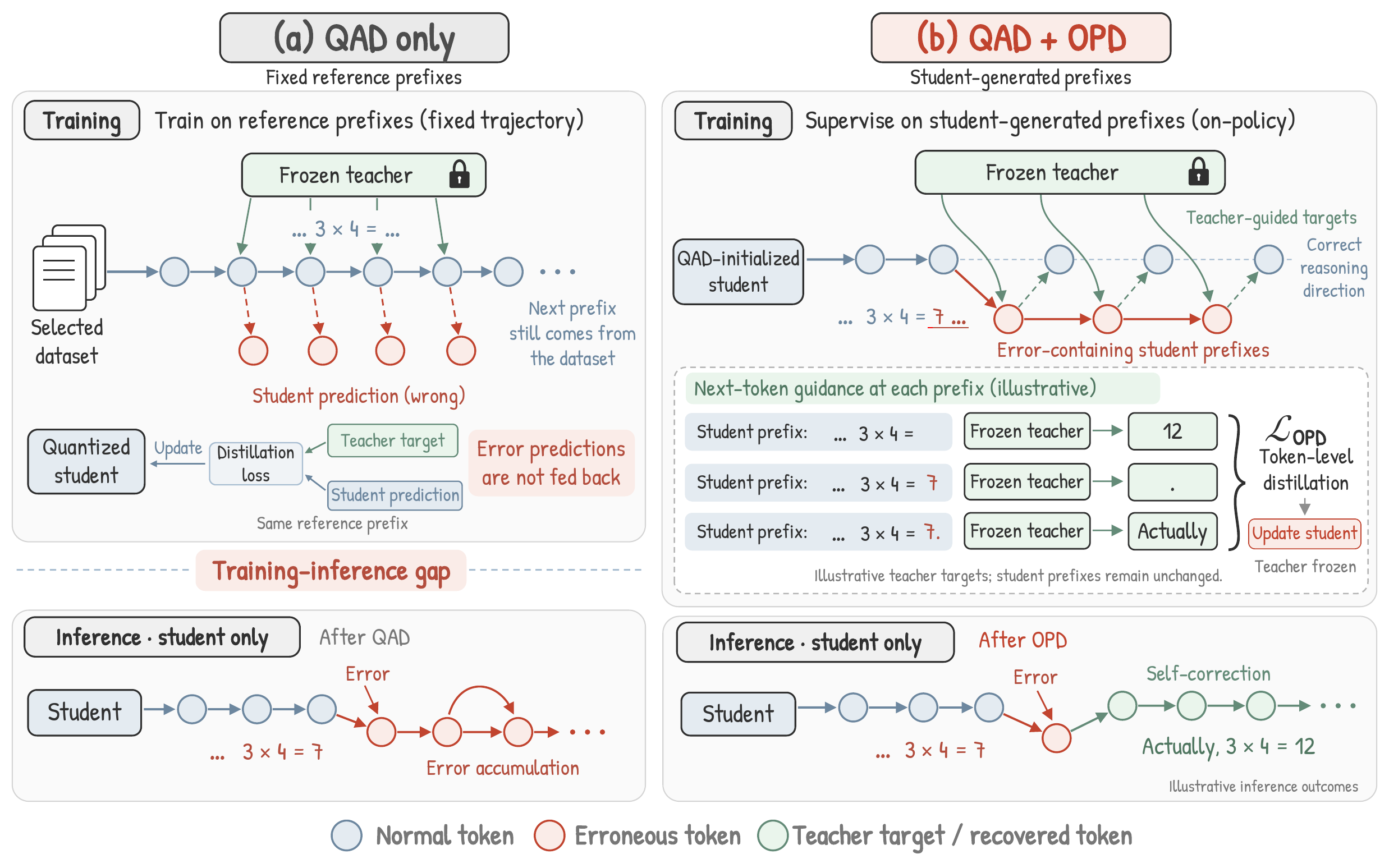}
\vspace{-12pt}
\caption{Two-stage recovery below three bits. QAD provides a stable low-bit
initialization through teacher supervision on fixed corpus prefixes, but that
supervision leaves a training--inference gap on
reasoning: the model is never trained on the trajectories it actually
generates at deployment. OPD closes this gap by sampling through the
deployment quantized path and combining token-level teacher guidance on
student prefixes with verifier rewards.}
\label{fig:overview}
\vspace{-12pt}
\end{figure}

\section{Related work}

\subsection{Extreme low-bit quantization and reasoning recovery}

Post-training quantization (PTQ) compresses pretrained models using a small
calibration set \citep{frantar2023gptq,shao2024omniquant}. Techniques such as
activation-aware scaling \mbox{\citep{xiao2023smoothquant,lin2024awq}}, rotations
\citep{ashkboos2024quarot}, mixed precision, and vector quantization
\citep{chee2023quip,egiazarian2024aqlm} reduce quantization error and preserve
performance at moderate precision. More aggressive compression motivates
quantization-aware training (QAT), which updates model parameters through a
simulated low-bit forward pass so that the model adapts to quantization during
training \citep{liu2024llmqat,ma2024bitnet}. Quantization-aware distillation
(QAD) augments this process with a full-precision teacher, transferring output
distributions or intermediate representations to recover the quantized
student's capabilities \citep{du2024bitdistiller}. In teacher-forced QAD, this
supervision is evaluated on prefixes drawn from a fixed corpus.

Recent studies examine the particular challenges of reasoning under extreme
quantization. \citet{lee2026reqat} identify how quantization errors
concentrated on low-entropy tokens propagate along a chain of thought.
\citet{lv2026reasoningqat} show that reinforcement learning applied to a
collapsed low-bit model requires a distillation cold start to establish a
viable policy.

\subsection{Exposure bias and on-policy distillation}

Exposure bias arises when a model is trained on reference prefixes but
conditions on its own predictions during autoregressive inference. Once
generation departs from the reference trajectory, subsequent predictions depend
on contexts that training may not have covered. Work in imitation learning
\citep{ross2011dagger} and sequence modeling addresses this mismatch by
exposing the learner to its own outputs, including through scheduled sampling
\citep{bengio2015scheduled} and sequence-level training
\citep{ranzato2016mixer}.

On-policy distillation brings teacher supervision directly onto these
student-generated trajectories \citep{gu2024minillm,agarwal2024gkd}. The
student samples completions, and the teacher supplies targets on the prefixes
the student actually visits. Relative to teacher-forced distillation
\citep{kim2016seqkd}, OPD changes the distribution of prefixes receiving
supervision \citep{ko2024distillm}. Relative to reward-only training, it provides
token-level teacher feedback throughout each completion. This combination
supports reasoning recovery by aligning dense supervision with the student's
own generation behavior. We adapt this idea, established in full-precision
post-training \citep{thinkingmachines2025opd}, to extreme low-bit
quantization.

\section{What teacher-forced recovery leaves behind}

\label{sec:diagnosis}

To understand what QAD recovers and what it leaves behind, we quantize
Qwen3-0.6B, Qwen3-1.7B, and Qwen3-4B using round-to-nearest (RTN) at effective
weight precisions of 2.79 and 1.88 bits, with 8-bit activations and 4-bit
embedding and output-head weights. We then recover the resulting six quantized
checkpoints using the EdgeRazor QAD recipe \citep{zhang2026edgerazor}.
Evaluation covers mathematical
reasoning on GSM8K, MATH-500, and AMC23, code generation on HumanEval and MBPP,
and 9 short-form question-answering benchmarks (QA9), with retention defined as each
checkpoint's performance as a percentage of the corresponding BF16 reference.

RTN leaves these models with zero accuracy on all five generative benchmarks.
QAD restores useful behavior, but the recovery is strikingly uneven across
short answers and extended reasoning.

\subsection{Short answers recover, long derivations do not}

\begin{figure}[t]
\centering
\includegraphics[width=\linewidth]{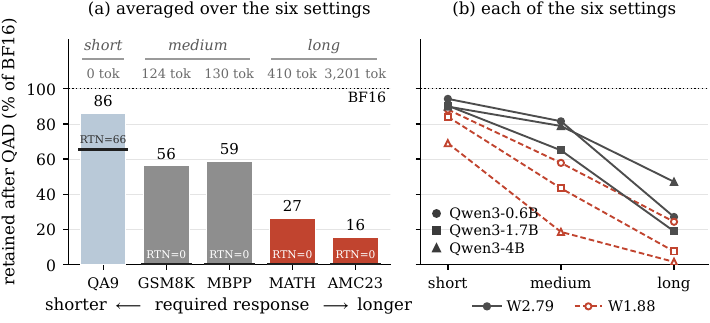}
\caption{QAD retention relative to BF16 across three Qwen3 scales and two
bit widths. (a) Benchmarks ordered by mean BF16 generation length
(shown above the bars); horizontal marks indicate RTN retention. QA9 uses
answer-option likelihoods without generation; MATH denotes MATH-500.
(b) Mean retention decreases from short- to medium- to long-form
tasks in every setting.}
\label{fig:dissociation}
\vspace{-8pt}
\end{figure}

The performance recovered by QAD separates into three distinct bands in
Figure~\ref{fig:dissociation}. Averaged over the six settings, the QAD
checkpoints retain 86\% of BF16 performance on QA9. Retention falls to 56\% on GSM8K and 59\% on MBPP,
then drops further to 27\% on MATH-500 and 16\% on AMC23.

What distinguishes these bands is how much the model must generate for itself.
QA9 scores answer options by likelihood and requires no autoregressive
generation. The BF16 reference generates roughly 130 tokens on GSM8K and MBPP,
compared with about 410 on MATH-500 and 3,200 on AMC23. GSM8K and MBPP require
different outputs, an arithmetic derivation and a working program, yet their
comparable generation lengths are accompanied by similar retention.

Figure~\ref{fig:dissociation}(b) shows the same decline from short to medium to
long responses in every model--bit-width setting. The recovery gap therefore
grows where the student must sustain a longer chain of its own predictions.

\subsection{Quantization amplifies exposure bias}

This dependence on the student's own predictions exposes a mismatch in
teacher-forced QAD. Its teacher supervision is evaluated on fixed prefixes
taken from the training corpus, while deployment visits prefixes the quantized
student generates for itself. Quantization amplifies this mismatch by changing
the trajectories the student follows.

At each generation step, quantization perturbs the next-token distribution.
Once this changes the selected token, every subsequent prediction conditions on
a different prefix. The continuation need not be incorrect: even a plausible
alternative can carry the student beyond the trajectories covered by teacher
forcing. Further predictions then combine quantization error with the effects
of that altered context, allowing deviations to accumulate along the sequence.
Lower precision increases the disruption, and longer generations provide more
opportunities for it to compound.

This mechanism predicts a breakdown that unfolds during generation: as
deviations accumulate, responses become harder to terminate and increasingly
prone to repetition.

\subsection{QAD loses the trajectory, not the arithmetic}

\label{sec:degeneration}

\begin{figure}[t]
\centering
\includegraphics[width=\linewidth]{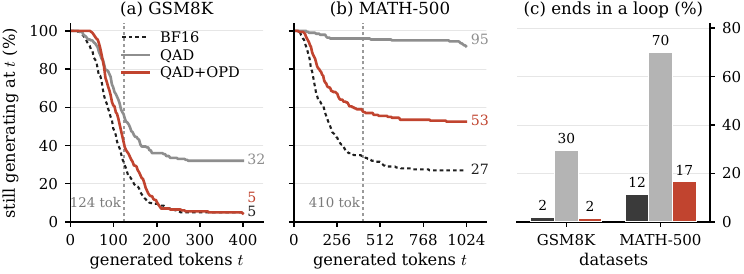}
\caption{OPD restores trajectory control after QAD. Qwen3-0.6B at W2.79
is evaluated with greedy decoding on 200 problems per benchmark.
(a, b) Fraction of responses still generating after $t$ tokens on GSM8K
and MATH-500. Vertical dashed lines mark mean BF16 response lengths;
endpoint labels report budget exhaustion rates.
(c) Fraction of responses ending in repeated 8-grams.}
\label{fig:method}
\vspace{-8pt}
\end{figure}

We track these two behaviors in the BF16 and W2.79 QAD checkpoints of
Qwen3-0.6B, using greedy generation on 200 problems each from GSM8K and
MATH-500. Figure~\ref{fig:method}(a,~b) shows the fraction of responses still
generating after $t$ tokens, while Figure~\ref{fig:method}(c) shows the
fraction whose closing words repeat an 8-gram.

The clearest breakdown appears on MATH-500: 95\% of QAD generations exhaust the
decoding budget, compared with 27\% for BF16. Moreover, 70\% of QAD outputs end
in repeated 8-grams, against 12\% for BF16. The model continues producing
tokens, but its reasoning becomes trapped in repetitions that prevent it from
reaching a conclusion.

The curves reveal how this gap develops. BF16 and QAD have similar termination
patterns over the first few dozen tokens, but their curves separate as
generation continues. Shorter GSM8K responses follow the same pattern with
smaller gaps: 32\% of QAD responses exhaust the budget, versus 5\% for BF16,
while repetition rates are 30\% and 2\%. The longer reasoning required by
MATH-500 exposes a much greater breakdown in the ability to sustain and
complete a solution.

These failed generations make the supervision gap concrete. QAD teaches the
student how to continue demonstrated trajectories, but the model must complete
its reasoning from prefixes produced by its own perturbed predictions.
Recovering this missing ability calls for extending teacher supervision to the
reasoning trajectories the quantized model actually generates.

\section{On-policy distillation for low-bit reasoning}

\label{sec:method}

Long-form reasoning breaks down along the quantized model's own trajectories;
recovering it calls for teacher guidance along those same paths. We therefore
combine QAD initialization with on-policy distillation. QAD first restores
broad capabilities and establishes a viable low-bit policy, providing the
foundation for rollout-based recovery: a
round-to-nearest model at these widths has nothing worth sampling and earns no
reward to bootstrap from. OPD then trains the student on trajectories sampled
through its deployment quantized forward path, with a frozen BF16 teacher
guiding its continuations and task verifiers rewarding successful solutions.
The two stages divide the work of recovery: QAD restores policy viability; OPD
restores trajectory control.

\subsection{From teacher forcing to student forcing}

Let $\pT$ denote a frozen BF16 teacher and $\pS$ the student policy evaluated
through the quantized forward path used at deployment. Teacher-forced QAD
learns from a corpus $\mathcal{D}$ of prompt--response pairs by minimizing

{\setlength{\abovedisplayskip}{5pt}\setlength{\belowdisplayskip}{5pt}
\begin{equation}
\mathcal{L}_{\mathrm{TF}}(\theta)
= \mathbb{E}_{(x,y)\sim\mathcal{D}}\!\left[\sum_{t=1}^{|y|}
\mathrm{KL}\!\left(\pT(\cdot\mid x,y_{<t})\,\middle\|\,\pS(\cdot\mid x,y_{<t})\right)
\right].
\label{eq:tf}
\end{equation}}

Here, each prefix $y_{<t}$ comes from a fixed reference response. Student
forcing instead draws prompts $x$ from a set $\mathcal{X}$ and samples a
completion $\hat{y}\sim\pS(\cdot\mid x)$, placing teacher supervision on the
resulting student-generated prefixes:

{\setlength{\abovedisplayskip}{5pt}\setlength{\belowdisplayskip}{5pt}
\begin{equation}
\mathcal{L}_{\mathrm{SF}}(\theta)
= \mathbb{E}_{x\sim\mathcal{X}}\;\mathbb{E}_{\hat{y}\sim\pS(\cdot\mid x)}
\!\left[\sum_{t=1}^{|\hat{y}|}
\mathrm{KL}\!\left(\pS(\cdot\mid x,\hat{y}_{<t})\,\middle\|\,\pT(\cdot\mid x,\hat{y}_{<t})\right)
\right].
\label{eq:sf}
\end{equation}}

Sampling through the quantized forward path makes the effects of low-bit
computation part of the training distribution itself. The teacher therefore
guides the student on continuations shaped by quantization, including
departures from the reference trajectories. We use reverse KL because it can be
estimated from student samples and penalizes continuations that the student
favors but the teacher assigns low probability to.

\subsection{Learning from quantized rollouts}

To translate this supervision into successful reasoning, we combine on-policy
teacher feedback with task-verifier rewards. For each prompt $x\in\mathcal{X}$,
the student generates $G$ completions through the quantized forward path. The
frozen teacher scores those same token sequences, and a verifier assigns each
completion a reward $r(x,\hat{y})$: final-answer correctness for mathematics
and test execution for code. We optimize

{\setlength{\abovedisplayskip}{5pt}\setlength{\belowdisplayskip}{5pt}
\begin{equation}
\mathcal{L}_{\mathrm{OPD}}(\theta)
= -\,\mathbb{E}\!\left[\sum_{t}\hat{A}(x,\hat{y})\,
\log\pS(\hat{y}_t\mid x,\hat{y}_{<t})\right]
\;+\;\beta\,\mathbb{E}\!\left[\sum_{t}
\log\frac{\pS(\hat{y}_t\mid x,\hat{y}_{<t})}{\pT(\hat{y}_t\mid x,\hat{y}_{<t})}\right],
\label{eq:opd}
\end{equation}}

where $\hat{A}$ is the advantage of $r(x,\hat{y})$ relative to the other
completions for the same prompt \citep{shao2024deepseekmath}, both expectations
are over $x\sim\mathcal{X}$ and $\hat{y}\sim\pS(\cdot\mid x)$, and $\beta$
controls the strength of teacher guidance. We use $\beta=1$ in every run
reported here, so the balance between the two terms is never tuned per
setting. The two signals guide recovery at
complementary scales: the sampled reverse-KL term supplies token-level feedback
on the prefixes the student actually generates, while the group-relative
advantage promotes completions that reach verified solutions.

\subsection{OPD restores trajectory control}

Training on these trajectories directly improves the generation behaviors that
break down after QAD. For the Qwen3-0.6B W2.79 checkpoint examined above,
Figure~\ref{fig:method}(a--c) shows the clearest recovery on MATH-500. OPD
reduces the loop rate from 70\% to 17\%, close to the BF16 rate of 12\%, and
cuts budget exhaustion from 95\% to 53\%. Accuracy rises alongside this
behavioral recovery, from 10.4\% to 24.2\%, approaching the BF16 reference of
27.2\%.

On GSM8K, the loop rate falls from 30\% to 2\% and budget exhaustion from 32\%
to 5\%, matching the BF16 reference on both measures. The larger improvement on
MATH-500 mirrors the greater disruption on longer generations: OPD recovers
more ground where QAD's trajectory failures are most severe. By extending
teacher guidance onto the student's own prefixes, OPD helps the low-bit model
sustain a derivation and bring it to a conclusion.

\section{Experiments}

\label{sec:experiments}

We evaluate reasoning recovery across four models and two effective bit widths,
examining the preservation of broad capabilities, the contribution of OPD at
lower precision, OPD's recovery efficiency relative to QAD, and how the
recovered models compare with existing methods.

\subsection{Experimental setup}

\label{sec:setup}

\begin{table}[b]
\vspace{-10pt}
\caption{OPD phases. The code phase resumes from the selected mathematics
checkpoint. Each optimizer step draws $8$ prompts and $G$ completions.
GSM8K, MATH L1--3, and MBPP are taken from the training splits and do not
overlap the evaluation sets.}
\label{tab:opd}
\centering
\small
\renewcommand{\arraystretch}{1.15}
\begin{tabular*}{\linewidth}{@{\extracolsep{\fill}}llcccc@{}}
\toprule
Phase & Training corpus & Steps & LR & $G$ & Rollouts \\
\midrule
Math & GSM8K, MATH L1--3, DAPO-Math & $80$--$140$
  & $3\times10^{-6}$ & $4$ & $32$ \\
Code & MBPP, KodCode & $80$--$150$
  & $3\times10^{-6}$ & $8$ & $64$ \\
\bottomrule
\end{tabular*}
\vspace{-8pt}
\end{table}

\paragraph{Models and quantization.} We evaluate Qwen3-0.6B, 1.7B, and 4B
\citep{qwen3}, together with Falcon3-1B-Instruct \citep{falcon3}, spanning
model scales and architectures. Following \citet{zhang2026edgerazor},
we use 2.79 and 1.88 effective bits per weight, allocating 50\% and 12.5\% of
weight groups to 4 bits, respectively, and the remainder to 1.58 bits.
Embedding and output-head weights use 4 bits and activations 8 bits.
Appendix~\ref{sec:edgerazor} details how we reproduced the QAD stage.

\paragraph{Implementation.} Every OPD run starts from the reported QAD
checkpoint, which keeps the QAD and OPD numbers measured from a common
starting point. Recovery then
proceeds in two phases, a mathematics phase followed by a code phase that
resumes from the selected mathematics checkpoint; Table~\ref{tab:opd} lists the
corpora, step budgets, and rollout settings, including the DAPO-Math
\mbox{\citep{yu2025dapo}} and KodCode \citep{xu2025kodcode} training sets. We train
with verl \citep{sheng2024verl} and generate with vLLM \citep{kwon2023vllm},
using non-thinking mode at temperature 1. The two precisions stay separate
within a step: BF16 master weights receive the updates, while rollouts pass
through the deployment quantized forward path. Each student is supervised by
its own BF16 counterpart as the frozen teacher, with a shared tokenizer;
Qwen3-0.6B is the one exception and takes Qwen3-1.7B.
Appendix~\ref{sec:teacher} examines teacher choice.
The learning rate is $3\times10^{-6}$, except for the
Falcon3 code phase, which uses $1\times10^{-6}$. We checkpoint every 20 steps
during OPD and select by validation on the current phase's metric.
Appendix~\ref{sec:curves} plots the phase-1 curves as an example.

\begin{table}[t]
\caption{Four models at two effective bit widths, from the BF16 reference down
to RTN and back up through QAD and OPD; the shaded row is the OPD result for
each line.}
\label{tab:main}
\centering
\footnotesize
\setlength{\tabcolsep}{2.5pt}
\renewcommand{\arraystretch}{1.05}
\begin{tabularx}{\linewidth}{l@{\hspace{4pt}}l@{\hspace{4pt}}lCCCCCC}
\toprule
& & & \mbox{GSM8K} & \mbox{MATH-500} & \mbox{AMC23} & \mbox{MBPP}
  & \mbox{HumanEval} & \mbox{QA9 (avg)} \\
\midrule
\multirow{7}{*}{Qwen3-0.6B} & \multicolumn{2}{l}{BF16} & 41.62 & 27.20 & 7.81 & 40.0 & 36.6 & 46.08 \\
\cmidrule{2-9}
 & W2.79 & RTN & 0.00 & 0.00 & 0.00 & 0.0 & 0.0 & 34.87 \\
 & & QAD & 32.75 & 10.40 & 1.25 & 33.7 & 29.9 & 43.43 \\
  &  & \oc$+$\,OPD & \oc\best{43.14} & \oc\best{24.20} & \oc\best{4.69} & \oc\best{37.7} & \oc\best{35.4} & \oc\best{43.64} \\
 & W1.88 & RTN & 0.00 & 0.00 & 0.00 & 0.0 & 0.0 & 34.74 \\
 & & QAD & 23.05 & 2.40 & 3.12 & 24.1 & 22.0 & 40.62 \\
  &  & \oc$+$\,OPD & \oc\best{37.65} & \oc\best{15.00} & \oc3.12 & \oc\best{35.7} & \oc\best{35.4} & \oc\best{41.83} \\
\midrule
\multirow{7}{*}{Qwen3-1.7B} & \multicolumn{2}{l}{BF16} & 68.76 & 54.40 & 31.72 & 54.0 & 67.1 & 54.54 \\
\cmidrule{2-9}
 & W2.79 & RTN & 0.00 & 0.00 & 0.00 & 0.0 & 0.0 & 35.49 \\
 & & QAD & 44.50 & 19.60 & 0.62 & 35.3 & 41.5 & 49.28 \\
  &  & \oc$+$\,OPD & \oc\best{54.28} & \oc\best{45.60} & \oc\best{19.38} & \oc\best{52.0} & \oc\best{59.1} & \oc\best{51.81} \\
 & W1.88 & RTN & 0.00 & 0.00 & 0.00 & 0.0 & 0.0 & 35.22 \\
 & & QAD & 26.46 & 8.20 & 0.00 & 26.1 & 24.4 & 45.85 \\
  &  & \oc$+$\,OPD & \oc\best{44.05} & \oc\best{24.20} & \oc\best{3.12} & \oc\best{41.3} & \oc\best{45.1} & \oc\best{47.90} \\
\midrule
\multirow{7}{*}{Qwen3-4B} & \multicolumn{2}{l}{BF16} & 86.35 & 68.80 & 55.00 & 67.6 & 82.3 & 61.63 \\
\cmidrule{2-9}
 & W2.79 & RTN & 0.00 & 0.00 & 0.00 & 0.0 & 0.0 & 34.80 \\
 & & QAD & 69.45 & 39.80 & 20.00 & 52.2 & 62.2 & 55.38 \\
  &  & \oc$+$\,OPD & \oc\best{77.86} & \oc\best{54.40} & \oc\best{39.38} & \oc\best{57.8} & \oc\best{64.0} & \oc\best{57.37} \\
 & W1.88 & RTN & 0.00 & 0.00 & 0.00 & 0.0 & 0.0 & 34.78 \\
 & & QAD & 17.36 & 2.20 & 0.00 & 11.6 & 11.0 & 42.56 \\
  &  & \oc$+$\,OPD & \oc\best{64.59} & \oc\best{36.00} & \oc\best{16.25} & \oc\best{48.7} & \oc\best{60.4} & \oc\best{52.83} \\
\midrule
\multirow{7}{*}{\shortstack[l]{Falcon3-1B\\-Instruct}}
 & \multicolumn{2}{l}{BF16} & 41.55 & 23.00 & 6.25 & 26.8 & 17.7 & 54.17 \\
\cmidrule{2-9}
 & W2.79 & RTN & 0.00 & 0.40 & 0.00 & 0.0 & 0.0 & 37.74 \\
 & & QAD & 36.32 & 16.00 & 5.00 & \best{24.6} & 20.7 & 52.89 \\
  &  & \oc$+$\,OPD & \oc\best{38.89} & \oc\best{19.20} & \oc2.50 & \oc24.1 & \oc20.7 & \oc\best{53.03} \\
 & W1.88 & RTN & 0.00 & 0.60 & 0.00 & 0.0 & 0.0 & 35.36 \\
 & & QAD & 31.61 & 12.40 & 2.50 & 21.0 & 14.6 & 49.53 \\
  &  & \oc$+$\,OPD & \oc\best{35.03} & \oc\best{16.20} & \oc\best{6.25} & \oc\best{24.3} & \oc\best{20.1} & \oc\best{49.58} \\
\bottomrule
\end{tabularx}
\vspace{-8pt}
\end{table}

\paragraph{Evaluation.} GSM8K \citep{cobbe2021gsm8k} and MATH-500
\citep{lightman2024letsverify} follow the lm-eval
protocol \citep{lmevalharness}. AMC23 uses avg@16 over 40 problems. Code
generation is measured by pass@1 on 448 MBPP problems \citep{austin2021mbpp}
and 164 HumanEval problems \citep{chen2021humaneval}. QA9 averages ARC-Easy,
ARC-Challenge, HellaSwag, SocialIQA, OpenBookQA, PIQA,
WinoGrande, TruthfulQA and MMLU, with per-task scores in
Appendix~\ref{sec:qa9}.

\subsection{Recovering reasoning while preserving broad capabilities}

For each configuration, Table~\ref{tab:main} compares four arms: the BF16
reference, RTN, QAD, and QAD$+$OPD. Building on QAD initialization, OPD
substantially restores mathematical and
code reasoning. OPD doubles average BF16 retention on
MATH-500 from 35.3\% to 69.7\% and raises GSM8K retention from 62.7\% to
85.0\%. Across the six Qwen3 configurations, GSM8K accuracy improves by
8.41--47.23 percentage points.
Code generation improves alongside mathematical reasoning. At 2.79 bits,
Qwen3-1.7B raises MBPP from 35.3\% to 52.0\% and HumanEval from 41.5\%
to 59.1\%, approaching the corresponding BF16 scores of 54.0\% and 67.1\%.

OPD successfully preserves the broad capabilities restored by QAD, with mean
QA9 retention increasing from 88.1\% to 92.1\%, so the combined pipeline
restores reasoning while maintaining short-form performance. OPD's
contribution grows as precision falls. Across all four models, OPD
accounts for a larger share of total GSM8K recovery from RTN at 1.88 bits than
at 2.79 bits. For Qwen3-4B, this share rises from 11\% to 73\%: at 1.88 bits,
OPD raises GSM8K accuracy from 17.36\% to 64.59\% and MBPP from 11.6\%
to 48.7\%.

\subsection{Recovery efficiency}

\begin{wraptable}{l}{0.45\linewidth}
\vspace{-\intextsep}
\caption{Training cost of QAD initialization and OPD recovery.}
\label{tab:cost}
\centering
\small
\setlength{\tabcolsep}{4pt}
\begin{tabular}{@{}llccc@{}}
\toprule
Model & Stage & Steps & \shortstack{Wall-\\clock} & \shortstack{GPU-\\hours} \\
\midrule
\multirow{2}{*}{\shortstack[l]{Qwen3-4B\\(W$1.88$)}}
  & QAD & $6{,}400$ & $102$\,h & $820$ \\
  & \gc OPD & \gc$260$ & \gc$14$\,h & \gc$57$ \\
\midrule
\multirow{2}{*}{\shortstack[l]{Falcon3-1B\\(W$2.79$)}}
  & QAD & $6{,}400$ & $32$\,h & $253$ \\
  & \gc OPD & \gc$245$ & \gc$2.8$\,h & \gc$11$ \\
\bottomrule
\end{tabular}
\vspace{-8pt}
\end{wraptable}

OPD converts training steps into reasoning gains far more efficiently than
continued teacher-forced QAD, yielding more GSM8K points per thousand
optimizer steps from the same starting checkpoint by up to $42\times$ where
continued QAD still gains at all. This translates into substantially lower
cost. Table~\ref{tab:cost} reports the two stages from the training logs: the
OPD stage uses $57$ GPU-hours against $820$ for QAD on Qwen3-4B at $1.88$
bits, and $11$ against $253$ on Falcon3-1B at $2.79$ bits, completing recovery
with approximately $14$--$23\times$ fewer GPU-hours than QAD initialization.

\subsection{Comparison with quantization baselines}

\begin{figure}[t]
\centering
\includegraphics[width=\linewidth]{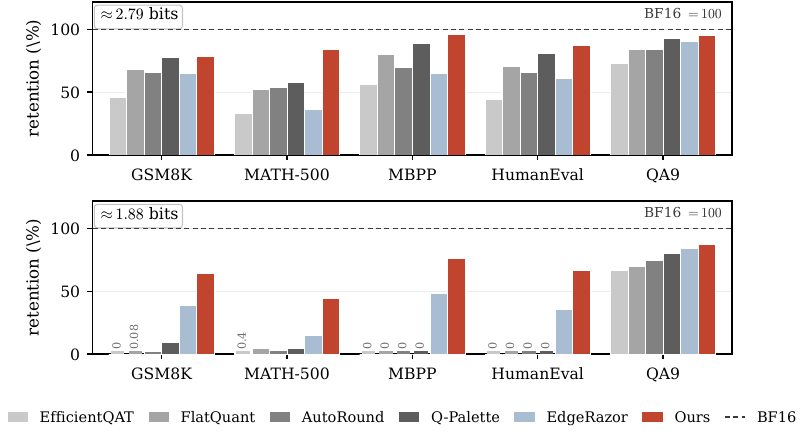}
\caption{Qwen3-1.7B as retention of the BF16 reference. Baselines are aligned to
our deployment setting of W$2.79$/W$1.88$ with INT8 activations, and to INT4
embedding and lm\_head. Methods without mixed-precision support fall back to the next whole
width, $3$ and $2$ bits in the two bands, which leaves them above our budget;
Appendix~\ref{sec:alignment} gives the builds.}
\label{fig:baselines}
\vspace{-8pt}
\end{figure}

The comparison in Figure~\ref{fig:baselines} places our pipeline on Qwen3-1.7B
against baselines spanning all three regimes of low-bit recovery: PTQ,
QAT, and QAD. Three PTQ methods cover signed-gradient
rounding \citep{cheng2024autoround}, learned affine transformations
\citep{sun2025flatquant}, and
fractional-bit codebook quantizers \citep{lee2025qpalette}, while a block-wise
method \citep{chen2024efficientqat} represents QAT, and the QAD arm is
EdgeRazor, which our own recovery also starts from.

At $2.79$ bits the ordering separates short-form from long-form ability. The
strongest baseline, Q-Palette, retains $93\%$ of BF16 on QA9 and $89\%$ on
MBPP, confirming that a well-designed quantizer preserves knowledge and short
code completions. Its retention falls to $58\%$ on MATH-500, while OPD reaches
$84\%$. The pattern repeats across the other baselines: on MATH-500 they retain
$33$--$58\%$, and none exceeds $59\%$, whereas OPD retains $84\%$ while running at $2.79$ bits.

Below two bits the baselines stop producing usable derivations altogether.
Every post-training and quantization-aware baseline scores zero on MBPP and
HumanEval, and at most $9\%$ retention on GSM8K, while still retaining
$67$--$80\%$ on QA9: the knowledge a likelihood-scored suite measures survives
a width at which derivations do not. EdgeRazor lifts the generative benchmarks
off zero, and OPD then multiplies what QAD recovers by $1.6$--$2.9\times$ on
each of them, reaching $64\%$ on GSM8K and $77\%$ on MBPP. Short-form
performance is unaffected by this shift: QA9 retention moves from $84\%$ to
$88\%$. The methods that compete with us on knowledge retention are therefore
not the ones that recover reasoning.

\section{Ablations}

\label{sec:ablations}

To quantify the benefit of switching to on-policy recovery after QAD
initialization, we compare OPD with continued teacher-forced QAD under matched
training conditions. Both arms use the same starting checkpoints, corpora,
mathematics-to-code schedule, and optimizer-step budgets. For continued QAD,
we generate reference responses with the same teacher on the training prompts.

Across Qwen3-0.6B and Qwen3-1.7B at both bit widths, OPD outperforms continued
QAD on all four benchmarks, winning all 16 comparisons
(Table~\ref{tab:control}). For Qwen3-1.7B at 2.79 bits, OPD nearly doubles
the MATH-500 accuracy of continued QAD, reaching 45.6\% versus 23.2\%. The same
configuration gains 9.70 GSM8K percentage points with OPD, while continued QAD
leaves accuracy essentially unchanged. More importantly, extending the QAD
mathematics phase to two and three times the matched budget yields essentially
no further gain across the benchmarks
(Appendix~\ref{sec:extended}). By contrast, switching to on-policy
training after QAD initialization unlocks reasoning gains that
continued teacher forcing fails to reach.

\begin{table}[t]
\caption{Controlled comparison of OPD and continued QAD under a matched
budget. Both arms share starting checkpoints, corpora, phase schedules,
step counts, learning rates, and samples per step.
OPD leads in all $16$ comparisons. Baselines are re-evaluated separately
for this control.}
\label{tab:control}
\centering
\footnotesize
\setlength{\tabcolsep}{2pt}
\renewcommand{\arraystretch}{1.05}
\begin{tabularx}{\linewidth}{l@{\hspace{5pt}}l@{\hspace{5pt}}lCCCC}
\toprule
Model & Width & Arm & \mbox{GSM8K} & \mbox{MATH-500} & \mbox{MBPP} & \mbox{HumanEval} \\
\midrule
\multirow{6}{*}{Qwen3-0.6B}
 & \multirow{3}{*}{W2.79} & QAD start & 33.21 & 10.00 & 33.7 & 29.9 \\
 & & $+$\,QAD, matched & 35.10 & 15.60 & 31.7 & 30.5 \\
 & & \gc$+$\,OPD & \gc\best{43.14} & \gc\best{24.20} & \gc\best{37.7} & \gc\best{35.4} \\
\cmidrule{2-7}
 & \multirow{3}{*}{W1.88} & QAD start & 22.52 & 2.60 & 24.1 & 22.0 \\
 & & $+$\,QAD, matched & 27.41 & 8.20 & 24.3 & 18.3 \\
 & & \gc$+$\,OPD & \gc\best{37.65} & \gc\best{15.00} & \gc\best{35.7} & \gc\best{35.4} \\
\midrule
\multirow{6}{*}{Qwen3-1.7B}
 & \multirow{3}{*}{W2.79} & QAD start & 44.58 & 20.60 & 35.3 & 41.5 \\
 & & $+$\,QAD, matched & 44.28 & 23.20 & 39.7 & 42.7 \\
 & & \gc$+$\,OPD & \gc\best{54.28} & \gc\best{45.60} & \gc\best{52.0} & \gc\best{59.1} \\
\cmidrule{2-7}
 & \multirow{3}{*}{W1.88} & QAD start & 26.54 & 8.40 & 26.1 & 24.4 \\
 & & $+$\,QAD, matched & 31.01 & 15.80 & 31.7 & 32.3 \\
 & & \gc$+$\,OPD & \gc\best{44.05} & \gc\best{24.20} & \gc\best{41.3} & \gc\best{45.1} \\
\bottomrule
\end{tabularx}
\vspace{-8pt}
\end{table}

\section{Conclusion}
\label{sec:conclusion}

Teacher-forced QAD restores broad short-form capabilities after extreme
quantization, yet leaves long reasoning trajectories vulnerable to accumulated
deviations and repetitive loops. OPD extends teacher supervision to prefixes
generated through the deployment quantized path, targeting the trajectories
where these failures arise. Across four models at 2.79 and 1.88 effective bits,
OPD doubles average BF16 retention on MATH-500 from 35\% to 70\%, improves code
generation, and preserves the broad capabilities restored by QAD. These gains
arrive in a few hundred optimizer steps, alongside improved termination and
reduced repetition.

\bibliography{refs}
\bibliographystyle{iclr2027_conference}

\appendix

\section{The EdgeRazor QAD initialization}
\label{sec:edgerazor}

Every arm in this paper starts from a checkpoint produced by EdgeRazor's
mixed-precision QAD recipe \citep{zhang2026edgerazor}, so what that stage does
and how faithfully we reproduced it bounds every number we report.

\paragraph{The quantizer.} Weights are quantized per block of $256$ input
channels. A block is taken either to ternary values $\{-1,0,1\}$, which costs
$\log_2 3 \approx 1.58$ bits, or to INT4. The ternary scale is not an absmax:
it is twice the block's mean absolute weight, so the rounding clips outliers
rather than stretching the grid to reach them, while INT4 blocks do use a
per-block absmax. Which blocks get the wider format is decided by position
alone. With a mixed-precision proportion $p$, rows of the output dimension are
selected at spacing $1/p$ and a selected row is INT4 across all of its blocks;
$p=0.5$ takes every second row and gives $0.5 \cdot 4 + 0.5 \cdot 1.58 = 2.79$
effective bits, and $p=0.125$ takes every eighth row for $1.88$. Because the
rule is positional rather than saliency-based, it needs no calibration data and
reproduces exactly. Embedding and output-head weights bypass the mixture and
are taken to per-block INT4. Activations on the decoder linears are INT8,
symmetric absmax over the same block size. Our arms leave the KV cache at $16$
bits, one step looser than the released \texttt{a8kv8} configuration; we note it
because the direction of that deviation favours us. Mathematics and QA are
evaluated on deployment exports of these weights; code is evaluated on the
trainable quantized checkpoint.

\paragraph{Our QAD run.} We reproduce the QAD stage ourselves, on the same
models, from the released repository and recipe, and every OPD arm in this
paper starts from a checkpoint of that reproduction. It distils on a general
instruction mixture, not on the mathematics and code pools OPD later uses, with
the student's own BF16 copy as teacher. The objective is an online logit KL
against that teacher plus a small task loss. We follow the released
hyperparameters -- learning rate
$2\times10^{-5}$ held constant after warmup, sequence length $1024$, 8-bit
AdamW, ZeRO-3 on eight GPUs -- with one deviation that matters and one we
judged not to. The effective batch is $768$, below the $1024$ and $1536$ the
recipe uses for Qwen3-0.6B and Qwen3-1.7B: the KL term materialises logits over
a $151{,}936$-token vocabulary, which at the published per-device batch exceeds
$70$\,GB on its own, so we reduced the
per-device batch and recovered part of it through gradient accumulation, taking
more optimizer steps inside a fixed token budget instead of restoring the batch.
We also skipped the optional offline step that regenerates each assistant turn
with the teacher, because the released training code reads the original files
rather than the regenerated ones, and because the recipe's supervision is the
online KL either way.

Table~\ref{tab:qadbudget} gives the per-arm budgets of that reproduction.

\begin{table}[h]
\caption{QAD budgets for the six Qwen3 arms. Per-device batch times gradient
accumulation gives $96$ sequences per GPU per optimizer step, so the effective
batch is $768$ on eight GPUs throughout. Wall-clock is the full QAD run.}
\label{tab:qadbudget}
\centering
\small
\renewcommand{\arraystretch}{1.15}
\begin{tabular*}{\linewidth}{@{\extracolsep{\fill}}llccccc@{}}
\toprule
Model & Width & Steps & Epochs & \mbox{Per-device $\times$ accum}
  & \mbox{Effective batch} & \mbox{Wall-clock} \\
\midrule
\multirow{2}{*}{Qwen3-0.6B} & W2.79 & $12{,}800$ & $2.0$ & $12\times8$ & $768$ & $55$\,h \\
 & W1.88 & $12{,}800$ & $2.0$ & $12\times8$ & $768$ & $55$\,h \\
\midrule
\multirow{2}{*}{Qwen3-1.7B} & W2.79 & $6{,}400$ & $1.0$ & $6\times16$ & $768$ & $44$\,h \\
 & W1.88 & $12{,}800$ & $2.0$ & $6\times16$ & $768$ & $87$\,h \\
\midrule
\multirow{2}{*}{Qwen3-4B} & W2.79 & $9{,}600$ & $1.5$ & $6\times16$ & $768$ & $124$\,h \\
 & W1.88 & $6{,}400$ & $1.0$ & $6\times16$ & $768$ & $102$\,h \\
\bottomrule
\end{tabular*}
\vspace{-8pt}
\end{table}

\section{QA9 per task}
\label{sec:qa9}

Table~\ref{tab:qa9} expands the QA9 column of Table~\ref{tab:main} into its
nine tasks, for every model, width and recovery stage.
Table~\ref{tab:qa9base} does the same for the QA9 column of
Figure~\ref{fig:baselines}.

\begin{table}[h]
\caption{QA9 by task. Columns are ARC-Easy, ARC-Challenge, HellaSwag,
SocialIQA, OpenBookQA, PIQA, WinoGrande, TruthfulQA and MMLU, each scored under
the protocol of Section~\ref{sec:setup}; MMLU is its own $57$-subject aggregate
and enters the mean as one task. The last column is the equally weighted mean
and reproduces the QA9 column of Table~\ref{tab:main}.}
\label{tab:qa9}
\centering
\footnotesize
\setlength{\tabcolsep}{2.5pt}
\renewcommand{\arraystretch}{1.05}
\begin{tabular}{@{}ll*{10}{c}@{}}
\toprule
& & ARC-e & ARC-c & HellaS & SIQA & OBQA & PIQA & WinoG & TQA & MMLU & QA9 \\
\midrule
\multicolumn{12}{@{}l}{\textit{Qwen3-0.6B}} \\
\multicolumn{2}{l}{BF16} & 56.19 & 33.79 & 47.26 & 39.25 & 31.40 & 67.36 & 56.27 & 42.84 & 40.34 & 46.08 \\
W2.79 & RTN & 25.51 & 25.77 & 26.44 & 34.29 & 27.20 & 53.32 & 48.46 & 48.00 & 24.82 & 34.87 \\
W2.79 & QAD & 53.45 & 30.20 & 37.57 & 40.89 & 28.20 & 64.15 & 54.70 & 44.01 & 37.70 & 43.43 \\
W2.79 & \oc$+$\,OPD & \oc53.20 & \oc30.97 & \oc38.75 & \oc40.43 & \oc29.00 & \oc65.07 & \oc53.91 & \oc45.23 & \oc36.17 & \oc43.64 \\
W1.88 & RTN & 25.97 & 26.28 & 26.16 & 33.16 & 27.20 & 50.92 & 48.78 & 49.65 & 24.56 & 34.74 \\
W1.88 & QAD & 47.90 & 28.07 & 33.53 & 37.67 & 28.00 & 63.06 & 52.49 & 43.98 & 30.89 & 40.62 \\
W1.88 & \oc$+$\,OPD & \oc50.63 & \oc28.67 & \oc35.53 & \oc40.43 & \oc27.20 & \oc64.31 & \oc53.75 & \oc41.31 & \oc34.65 & \oc41.83 \\
\midrule
\multicolumn{12}{@{}l}{\textit{Qwen3-1.7B}} \\
\multicolumn{2}{l}{BF16} & 70.20 & 43.43 & 60.37 & 45.09 & 37.40 & 72.25 & 60.85 & 45.84 & 55.45 & 54.54 \\
W2.79 & RTN & 26.05 & 25.00 & 26.55 & 33.52 & 29.60 & 52.39 & 50.59 & 49.84 & 25.85 & 35.49 \\
W2.79 & QAD & 63.80 & 37.12 & 49.65 & 43.65 & 32.20 & 69.15 & 54.54 & 48.84 & 44.54 & 49.28 \\
W2.79 & \oc$+$\,OPD & \oc68.98 & \oc41.21 & \oc56.77 & \oc48.26 & \oc33.00 & \oc71.33 & \oc59.27 & \oc44.30 & \oc43.21 & \oc51.81 \\
W1.88 & RTN & 26.30 & 25.94 & 26.10 & 34.08 & 28.20 & 51.69 & 48.62 & 50.23 & 25.86 & 35.22 \\
W1.88 & QAD & 58.00 & 32.94 & 44.41 & 41.66 & 31.00 & 66.21 & 55.41 & 43.44 & 39.56 & 45.85 \\
W1.88 & \oc$+$\,OPD & \oc61.99 & \oc36.86 & \oc46.59 & \oc43.81 & \oc34.00 & \oc68.61 & \oc54.22 & \oc44.64 & \oc40.34 & \oc47.90 \\
\midrule
\multicolumn{12}{@{}l}{\textit{Qwen3-4B}} \\
\multicolumn{2}{l}{BF16} & 78.58 & 53.75 & 68.40 & 49.95 & 40.40 & 75.03 & 65.59 & 54.68 & 68.30 & 61.63 \\
W2.79 & RTN & 25.46 & 25.26 & 26.43 & 33.21 & 27.60 & 51.36 & 50.83 & 49.07 & 24.02 & 34.80 \\
W2.79 & QAD & 73.53 & 45.73 & 57.83 & 47.13 & 37.00 & 72.58 & 61.56 & 51.16 & 51.93 & 55.38 \\
W2.79 & \oc$+$\,OPD & \oc75.59 & \oc49.23 & \oc60.08 & \oc48.52 & \oc38.00 & \oc74.05 & \oc64.72 & \oc49.62 & \oc56.54 & \oc57.37 \\
W1.88 & RTN & 24.96 & 26.02 & 26.57 & 33.93 & 29.00 & 50.38 & 48.70 & 49.34 & 24.08 & 34.78 \\
W1.88 & QAD & 45.41 & 28.92 & 37.41 & 42.63 & 28.80 & 64.53 & 55.09 & 44.08 & 36.16 & 42.56 \\
W1.88 & \oc$+$\,OPD & \oc70.79 & \oc44.28 & \oc52.76 & \oc47.95 & \oc33.00 & \oc72.52 & \oc59.91 & \oc49.52 & \oc44.74 & \oc52.83 \\
\midrule
\multicolumn{12}{@{}l}{\textit{Falcon3-1B-Instruct}} \\
\multicolumn{2}{l}{BF16} & 68.18 & 45.56 & 63.10 & 45.60 & 40.40 & 74.92 & 60.30 & 45.59 & 43.85 & 54.17 \\
W2.79 & RTN & 37.88 & 25.85 & 35.80 & 35.57 & 25.60 & 58.49 & 50.59 & 45.59 & 24.25 & 37.74 \\
W2.79 & QAD & 70.33 & 44.03 & 57.97 & 49.18 & 38.00 & 73.72 & 59.67 & 42.02 & 41.13 & 52.89 \\
W2.79 & \oc$+$\,OPD & \oc71.13 & \oc44.71 & \oc57.65 & \oc50.05 & \oc37.80 & \oc73.07 & \oc59.51 & \oc42.05 & \oc41.28 & \oc53.03 \\
W1.88 & RTN & 30.56 & 22.53 & 26.69 & 34.03 & 27.60 & 52.50 & 51.62 & 48.99 & 23.71 & 35.36 \\
W1.88 & QAD & 66.71 & 39.33 & 51.55 & 47.54 & 34.60 & 72.09 & 56.99 & 40.28 & 36.72 & 49.53 \\
W1.88 & \oc$+$\,OPD & \oc66.75 & \oc39.59 & \oc51.35 & \oc47.44 & \oc34.80 & \oc72.31 & \oc57.30 & \oc40.20 & \oc36.51 & \oc49.58 \\
\bottomrule
\end{tabular}
\vspace{-8pt}
\end{table}

\begin{table}[h]
\caption{QA9 by task for the Qwen3-1.7B arms of Figure~\ref{fig:baselines}.
Columns are as in Table~\ref{tab:qa9}; Ret.\ is the mean as a percentage of the
BF16 row and is what the figure plots. The BF16 row is the same unquantized
reference as in Table~\ref{tab:qa9}, and the EdgeRazor and Ours rows are that
table's QAD and $+$\,OPD rows for this model.}
\label{tab:qa9base}
\centering
\footnotesize
\setlength{\tabcolsep}{2pt}
\renewcommand{\arraystretch}{1.05}
\begin{tabular}{@{}ll*{11}{c}@{}}
\toprule
& & ARC-e & ARC-c & HellaS & SIQA & OBQA & PIQA & WinoG & TQA & MMLU & QA9 & Ret. \\
\midrule
\multicolumn{2}{l}{BF16} & 70.20 & 43.43 & 60.37 & 45.09 & 37.40 & 72.25 & 60.85 & 45.84 & 55.45 & 54.54 & --- \\
\midrule
\multicolumn{13}{@{}l}{\textit{$\approx 2.79$ bits}} \\
& EfficientQAT & 40.32 & 27.99 & 42.90 & 37.31 & 26.60 & 62.84 & 51.30 & 45.80 & 24.88 & 39.99 & $73.3$ \\
& FlatQuant & 52.57 & 33.70 & 48.98 & 38.54 & 30.00 & 66.87 & 56.04 & 44.22 & 42.99 & 45.99 & $84.3$ \\
& AutoRound & 54.08 & 32.17 & 49.06 & 41.86 & 32.60 & 66.87 & 56.35 & 44.19 & 36.65 & 45.98 & $84.3$ \\
& Q-Palette & 59.39 & 39.33 & 55.77 & 41.56 & 36.20 & 69.70 & 59.04 & 43.69 & 51.69 & 50.71 & $93.0$ \\
& EdgeRazor & 63.80 & 37.12 & 49.65 & 43.65 & 32.20 & 69.15 & 54.54 & 48.84 & 44.54 & 49.28 & $90.4$ \\
& \oc Ours & \oc68.98 & \oc41.21 & \oc56.77 & \oc48.26 & \oc33.00 & \oc71.33 & \oc59.27 & \oc44.30 & \oc43.21 & \oc51.81 & \oc$\mathbf{95.0}$ \\
\midrule
\multicolumn{13}{@{}l}{\textit{$\approx 1.88$ bits}} \\
& EfficientQAT & 33.75 & 23.81 & 33.09 & 35.06 & 25.80 & 58.87 & 50.67 & 45.29 & 22.95 & 36.59 & $67.1$ \\
& FlatQuant & 39.39 & 25.09 & 34.28 & 35.41 & 26.40 & 58.81 & 52.96 & 47.02 & 23.14 & 38.06 & $69.8$ \\
& AutoRound & 44.78 & 27.22 & 40.03 & 38.64 & 30.20 & 61.48 & 55.49 & 41.56 & 29.11 & 40.94 & $75.1$ \\
& Q-Palette & 48.65 & 31.31 & 45.45 & 39.92 & 31.60 & 64.91 & 54.22 & 42.31 & 35.42 & 43.76 & $80.2$ \\
& EdgeRazor & 58.00 & 32.94 & 44.41 & 41.66 & 31.00 & 66.21 & 55.41 & 43.44 & 39.56 & 45.85 & $84.1$ \\
& \oc Ours & \oc61.99 & \oc36.86 & \oc46.59 & \oc43.81 & \oc34.00 & \oc68.61 & \oc54.22 & \oc44.64 & \oc40.34 & \oc47.90 & \oc$\mathbf{87.8}$ \\
\bottomrule
\end{tabular}
\vspace{-8pt}
\end{table}

\section{Baseline alignment}
\label{sec:alignment}

Every baseline in Figure~\ref{fig:baselines} is brought onto the setting our own
arms run at, so that a score difference is the quantization method rather than a
difference in what was quantized.

\paragraph{Activations and embeddings.} All arms carry INT8 activations on the
decoder linears, symmetric absmax over blocks of $256$ input channels, the
quantizer our configuration uses. On QA9 the embedding and lm\_head are also
taken to INT4 over blocks of $256$, matching the override in our recipe; the
generative benchmarks leave them in BF16, which favours the baselines. The
isolated cost of the embedding override is small: on Qwen3-1.7B it moves GSM8K
by $-0.23$ points for BF16 weights and $-0.45$ for FlatQuant at $4$ bits.

\paragraph{Width.} Our arms mix INT1.58 and INT4 rows, $0.5$ of each, for
$2.79$ effective bits, and the analogous mixture for $1.88$. Q-Palette
constructs fractional widths natively and is run at the same two widths.
EfficientQAT, FlatQuant and AutoRound emit a single width per run, so they fall
back to the next whole width, $3$ bits against our $2.79$ and $2$ bits against
our $1.88$. Rounding up rather than down keeps the fallback in the baseline's
favour.

\paragraph{Width-matched QA9.} On QA9 the extra bit was large enough to
reorder the comparison, so the upper band also matches the width. Our row rule
is positional -- every second output row of every decoder linear is INT4 -- so
it can be reproduced without a saliency criterion. Lacking an INT1.58 output, a
baseline reaches $2.79$ from the widths it does produce, taking a fraction
$p$ of its $2$-bit solution and $1-p$ of a wider one: $p=0.21$ against $3$
bits, $p=0.605$ against $4$ bits. Both are built and the better one reported.
Mixing is row-wise for AutoRound and EfficientQAT, whose saved weights
approximate the original layer directly. FlatQuant folds a per-channel scale
into the preceding LayerNorm, so a row taken from one width would be expressed
in the other's basis; its mixture is therefore layer-wise, with each layer's
norms travelling with its linears.

\paragraph{Reproduction details.} Baselines run from their released
implementations, with one exception. Q-Palette's official entrypoints are
Llama-only and depend on custom CUDA kernels, so its arm is our own
pure-PyTorch reimplementation: a Walsh--Hadamard rotation, then output rows
split across two adjacent codebook levels so the mean width matches the target.
It is data-free, and it is also the arm that leads QA9 and MBPP, so the
comparison there is against our reading of the method rather than the authors'
code. The methods that do calibrate share one corpus, the $10$k-document Pile
subset that AutoRound uses by default, which we substituted for FlatQuant's
WikiText-2 and EfficientQAT's RedPajama defaults so that no arm differs from
another in what it saw; weight group size is $128$ wherever a method exposes
it, and sample counts and epoch budgets follow each method's published recipe.
One asymmetry is left uncorrected because it favours the baselines: on the
generative benchmarks their embedding and lm\_head stay in BF16 while ours are
INT4.

\section{Training efficiency}
\label{sec:efficiency}

\begin{figure}[h]
\centering
\includegraphics[width=\linewidth]{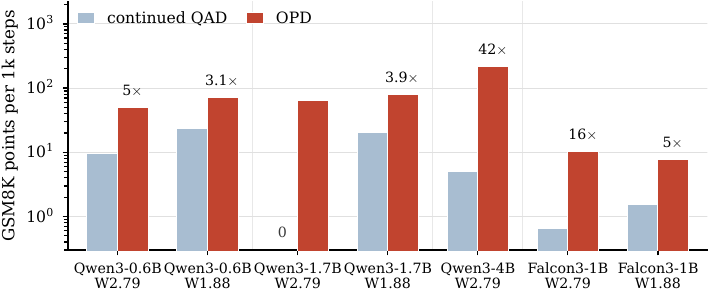}
\caption{OPD versus continued QAD on GSM8K, in points per thousand optimizer
steps from the same starting checkpoint. Qwen3-0.6B and Qwen3-1.7B use the
matched step budget of Table~\ref{tab:control}; Qwen3-4B and Falcon3-1B use a
late QAD segment. The vertical axis is logarithmic. A zero marks a
non-positive QAD gain.}
\label{fig:share}
\vspace{-8pt}
\end{figure}

Figure~\ref{fig:share} gives the per-step comparison behind
Section~\ref{sec:experiments}: across the four models, OPD yields more GSM8K
points per thousand optimizer steps than continued teacher-forced QAD from the
same starting checkpoint. For Qwen3-4B at 2.79 bits, just 20 OPD steps gain
4.32 points, exceeding the 4.09-point gain from 800 QAD steps.

\section{OPD training curves}
\label{sec:curves}

\begin{figure}[h]
\centering
\includegraphics[width=\linewidth]{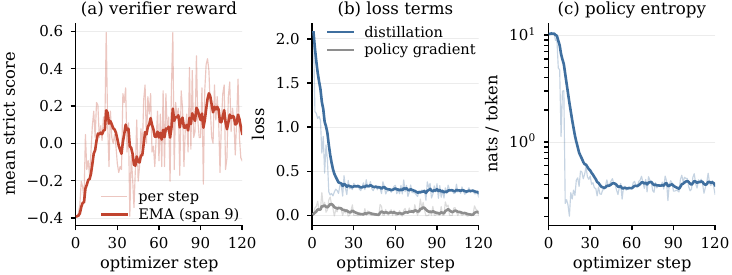}
\caption{OPD phase-1 training for Qwen3-4B at W$1.88$: $120$ optimizer steps
on the clean mathematics pool. Faint lines are per-step values and solid lines
an exponential moving average over nine steps. (a) mean strict verifier score
over the $32$ sequences of a step. (b) the two terms of
Equation~\ref{eq:opd}. (c) policy entropy, logarithmic axis.}
\label{fig:curves}
\vspace{-8pt}
\end{figure}

Figure~\ref{fig:curves} shows the phase-1 run for Qwen3-4B at $1.88$ bits, the
arm whose QAD start is the most degraded of the six and where the recovery is
therefore easiest to read. The run divides into a short repair phase and a
long plateau. Over the first thirty steps the policy entropy falls from about
$10$ nats per token, an effective support of tens of thousands of tokens, to
$0.35$: the quantized start is close to emitting arbitrary text, and teacher
supervision on its own prefixes restores a usable output distribution almost
immediately. The distillation term falls from $2.07$ to $0.25$ over the same
interval and the verifier reward rises from $-0.39$ to roughly $+0.15$. The
remaining ninety steps hold the reward between $0.1$ and $0.2$.

The two terms of Equation~\ref{eq:opd} sit on very different scales. With
$\beta=1$ the distillation term accounts for almost all of the total loss,
while the policy-gradient term stays near $0.05$ throughout. Single-step
reward spans roughly $\pm0.5$ because a step scores only $32$ sequences, so
the smoothed curve rather than the raw trace carries the trend.

\section{Teacher choice}
\label{sec:teacher}

A natural worry is that OPD's gain is borrowed capacity from a stronger
teacher rather than the on-policy objective. The default recipe uses each
student's own BF16 weights; Qwen3-0.6B is the one exception, because that
model's BF16 copy is too weak to supervise. Table~\ref{tab:teacher} separates
the two cases.

From $1.7$B up, a larger teacher does not help. On Qwen3-1.7B at $2.79$ bits,
self-distillation matches or beats $4$B and $8$B teachers on both GSM8K and
MATH-500, within run-to-run variation on GSM8K and ahead by $1.9$ points on
MATH-500. On Qwen3-4B at $2.79$ bits, switching the teacher to $8$B after the
self-teacher mathematics checkpoint does not improve either metric. At $1.88$
bits the same student's own BF16 weights are not merely sufficient but better:
matched $80$-step arms from the same QAD start gain $+18.88$ GSM8K with the
self teacher against $+15.62$ and $+15.84$ for the $4$B and $8$B teachers, and
lead MATH-500 by $6.4$ and $6.8$ points. On Qwen3-4B at $1.88$ bits the two
teachers are level instead: from a start that quantization had driven down to
$17.36$ GSM8K, the self teacher recovers $+48.60$ against $+47.77$ for the $8$B
teacher, and MATH-500 splits the other way, $+34.20$ against $+35.40$. Across
both widths and all three students, then, no teacher larger than the student
buys anything its own BF16 copy does not: the recovery reported in the main
text is the method, not a stronger model.

Qwen3-0.6B is a special case. A probe on the mathematics training pool, four
samples per problem, gives teacher pass@4 of $62\%$ for Qwen3-0.6B against
$83\%$ for $1.7$B and $89\%$ for both $4$B and $8$B, while the quantized
$0.6$B student sits at $36$--$39\%$. Its own BF16 teacher is barely ahead of
the student, so there is little to transfer. Matched OPD arms confirm the
gap: self-distillation gains only $+2.80$ and $+6.44$ GSM8K points at the two
widths, whereas every teacher of $1.7$B or above gains $+11.0$ to $+14.1$.
Scaling past $1.7$B buys nothing: at $2.79$ bits the $8$B teacher adds $0.23$
GSM8K points, below the spread between adjacent checkpoints of a single arm,
and at $1.88$ bits the $4$B and $8$B teachers fall $1.8$ and $3.1$ points
behind it, losing MATH-500 at both widths. We therefore keep Qwen3-1.7B as the
$0.6$B teacher and use every larger student's own BF16 weights.

\begin{table}[h]
\caption{Teacher size. Entries are gains over a QAD baseline fixed per student
and width, so a row reports what OPD recovered rather than what the model
scores. Baselines, as GSM8K / MATH-500: Qwen3-0.6B $33.21$ / $10.00$ at $2.79$
bits and $22.52$ / $2.60$ at $1.88$; Qwen3-1.7B $44.35$ / $17.80$ and $26.46$ /
$8.20$; Qwen3-4B $69.45$ / $39.80$ and $17.36$ / $2.20$. A baseline is shared
by every row of its block, so the differences between teachers do not depend on
it. Steps counts OPD steps, and both columns of a row come from the same
checkpoint. Every teacher within a block is trained to the same budget except
Qwen3-4B at $2.79$ bits, whose two rows are not matched. Shaded rows are the
teacher used in the main experiments.}
\label{tab:teacher}
\centering
\footnotesize
\renewcommand{\arraystretch}{1.12}
\begin{tabularx}{\linewidth}{CCCCCC}
\toprule
Student & Teacher & Width & Steps & $\Delta$GSM8K & $\Delta$MATH-500 \\
\midrule
 & & W2.79 & $80$ & $+2.80$ & $+9.40$ \\
 & \multirow{-2}{*}{self} & W1.88 & $90$ & $+6.44$ & $\mathbf{+9.20}$ \\
\cmidrule{2-6}
 & \gc & \gc W2.79 & \gc $80$ & \gc $+13.87$ & \gc $\mathbf{+13.80}$ \\
 & \gc \multirow{-2}{*}{$1.7$B} & \gc W1.88 & \gc $90$ & \gc $\mathbf{+14.10}$ & \gc $+9.00$ \\
\cmidrule{2-6}
 & & W2.79 & $80$ & $+11.75$ & $+10.00$ \\
 & \multirow{-2}{*}{$4$B} & W1.88 & $90$ & $+12.35$ & $+3.00$ \\
\cmidrule{2-6}
 & & W2.79 & $80$ & $\mathbf{+14.10}$ & $+11.00$ \\
\multirow{-8}{*}{Qwen3-0.6B} & \multirow{-2}{*}{$8$B} & W1.88 & $90$ & $+10.99$ & $+7.40$ \\
\midrule
 & \gc & \gc W2.79 & \gc $200$ & \gc $\mathbf{+9.20}$ & \gc $\mathbf{+13.55}$ \\
 & \gc \multirow{-2}{*}{self} & \gc W1.88 & \gc $80$ & \gc $\mathbf{+18.88}$ & \gc $\mathbf{+16.40}$ \\
\cmidrule{2-6}
 & & W2.79 & $200$ & $+8.77$ & $+11.65$ \\
 & \multirow{-2}{*}{$4$B} & W1.88 & $80$ & $+15.62$ & $+10.00$ \\
\cmidrule{2-6}
 & & W2.79 & $200$ & $+8.34$ & $+13.05$ \\
\multirow{-6}{*}{Qwen3-1.7B} & \multirow{-2}{*}{$8$B} & W1.88 & $80$ & $+15.84$ & $+9.60$ \\
\midrule
 & \gc & \gc W2.79 & \gc $80$ & \gc $\mathbf{+8.41}$ & \gc $\mathbf{+14.60}$ \\
 & \gc \multirow{-2}{*}{self} & \gc W1.88 & \gc $120$ & \gc $\mathbf{+48.60}$ & \gc $+34.20$ \\
\cmidrule{2-6}
 & & W2.79 & $60$ & $+7.20$ & $+13.40$ \\
\multirow{-4}{*}{Qwen3-4B} & \multirow{-2}{*}{$8$B} & W1.88 & $120$ & $+47.77$ & $\mathbf{+35.40}$ \\
\bottomrule
\end{tabularx}
\vspace{-8pt}
\end{table}

\clearpage
\section{Extended QAD budget}
\label{sec:extended}

The matched-budget control in Table~\ref{tab:control} equalizes optimizer
steps, so a remaining objection is that teacher-forced QAD simply needs a
longer run. We therefore keep the same starting checkpoints and mathematics
corpus, and extend the QAD mathematics phase to $2\times$ and $3\times$ the
matched step budget; the $3\times$ run covers roughly one epoch of the
training pool. Table~\ref{tab:extended} compares these arms with the OPD
mathematics phase alone.

Extra teacher-forced steps do not close the gap. At $3\times$, QAD still
trails OPD by $8.9$--$18.1$ GSM8K points on every configuration, and on
Qwen3-0.6B at $2.79$ bits the $3\times$ run falls below the $1\times$
baseline. The advantage of switching to on-policy recovery after QAD is
therefore not an artifact of the matched step count.

\begin{table}[h]
\caption{Mathematics-only recovery with QAD trained for $2\times$ and
$3\times$ the matched budget of Table~\ref{tab:control}. The OPD column
is the mathematics phase only; the gap is OPD minus QAD at $3\times$.
Baselines are re-evaluated separately for this control.}
\label{tab:extended}
\centering
\footnotesize
\setlength{\tabcolsep}{2pt}
\renewcommand{\arraystretch}{1.05}
\begin{tabularx}{\linewidth}{l@{\hspace{5pt}}l@{\hspace{5pt}}CCC>{\columncolor{ablrow}}CC}
\toprule
Model & Width & QAD $1\times$ & QAD $2\times$ & QAD $3\times$
  & \mbox{OPD phase 1} & gap \\
\midrule
\multirow{2}{*}{Qwen3-0.6B} & W2.79 & 33.43 & 36.39 & 27.67 & \best{45.72} & $+18.1$ \\
 & W1.88 & 30.63 & 26.16 & 27.29 & \best{37.23} & $+9.9$ \\
\midrule
\multirow{2}{*}{Qwen3-1.7B} & W2.79 & 45.34 & 48.75 & 44.05 & \best{52.92} & $+8.9$ \\
 & W1.88 & 34.87 & 38.06 & 36.32 & \best{45.34} & $+9.0$ \\
\bottomrule
\end{tabularx}
\vspace{-8pt}
\end{table}

\end{document}